%% file: llncs.tex
\documentclass{llncs}
\usepackage{makeidx}  
\usepackage{setspace}
\usepackage{amsmath,amsfonts}
\usepackage{microtype}
\usepackage{hyperref}
\usepackage{graphicx}
\usepackage{color}
\usepackage{bbm} 
\usepackage{ amssymb }

\newcommand{\bD}{\boldsymbol{D}}

\newcommand{\bL}{\boldsymbol{L}}

\newcommand{\bp}{\boldsymbol{p}}

\newcommand{\bA}{\boldsymbol{A}}
\newcommand{\ba}{\boldsymbol{a}}
\newcommand{\bb}{\boldsymbol{b}}

\newcommand{\bmu}{\boldsymbol{\mu}}

\newcommand{\bphi}{\boldsymbol{\phi}}
\newcommand{\bpsi}{\boldsymbol{\psi}}
\newcommand{\bSigma}{\boldsymbol{\Sigma}}

\newcommand{\bLambda}{\boldsymbol{\Lambda}}

\newcommand{\KL}{\text{KL}}

\newcommand{\by}{\boldsymbol{y}}
\newcommand{\bz}{\boldsymbol{z}}

\newcommand{\bv}{\boldsymbol{v}}

\newcommand{\bmoving}{\boldsymbol{x}}
\newcommand{\bfixed}{\boldsymbol{y}}

\newcommand{\moving}{x}
\newcommand{\fixed}{y}

\newcommand{\Expect}{{\rm I\kern-.3em E}}

\title{Unsupervised Learning for Fast \\ Probabilistic Diffeomorphic Registration}
\titlerunning{Generative Model for Diffeomorphic Registration}

\author{Adrian V. Dalca\inst{1,2,3} \and Guha Balakrishnan\inst{1} \and John Guttag\inst{1} \and Mert R. Sabuncu\inst{3} }
\authorrunning{Anon et al.}
\institute{Computer Science and Artificial Intelligence Lab, MIT \\
	\and Martinos Center for Biomedical Imaging, Massachusetts General Hospital, HMS \\
	\and School of Electrical and Computer Engineering, Cornell University}

\begin{document}
	   \abovedisplayskip=5pt
	   \belowdisplayskip=5pt

\pagestyle{headings}  
\maketitle              

\begin{abstract}
	\input{abstract}
\end{abstract}

\input{introduction}

\input{model}

\input{results}

\input{conclusion}
\vspace{-0.3cm}

\bibliographystyle{plain}
{\footnotesize
\bibliography{references}
}

\end{document}

%% file: abstract.tex

Traditional  deformable registration techniques achieve impressive results and offer a rigorous theoretical treatment, but are computationally intensive since they solve an optimization problem for each image pair. Recently, learning-based methods have facilitated fast registration by learning spatial deformation functions. 
However, these approaches use restricted deformation models, require supervised labels, or do not guarantee a diffeomorphic (topology-preserving) registration. Furthermore, learning-based registration tools have not been derived from a probabilistic framework that can offer uncertainty estimates.
In this paper, we present a probabilistic generative model and derive an unsupervised learning-based inference algorithm that makes use of recent developments in convolutional neural networks (CNNs). We demonstrate our method on a 3D brain registration task, and provide an empirical analysis of the algorithm. Our approach results in state of the art accuracy and very fast runtimes, while providing diffeomorphic guarantees and uncertainty estimates. Our implementation is available online at \url{https://github.com/voxelmorph/voxelmorph}.

\keywords{medical image registration, diffeomorphic registration, probabilistic modeling, convolutional neural networks, variational inference, uncertainty estimation}

%% file: introduction.tex
\vspace{-0.9cm}
\section{Introduction}
\label{sec:introduction}
\vspace{-0.3cm}

Deformable registration computes a dense correspondence between two images, and is fundamental to many medical image analysis tasks. 
%
%
Traditional methods  solve an optimization over the space of deformations, such as elastic-type models~\cite{bajcsy1989,shen2002}, B splines~\cite{rueckert1999}, dense vector fields~\cite{thirion1998}, or discrete methods~\cite{dalca2016,glocker2008}. Constraining the allowable transformations to diffeomorphisms ensures certain desirable properties, such as preservation of topology. Diffeomorphic transforms have seen extensive methodological development, yielding state-of-the-art tools, such as LDDMM~\cite{beg2005,zhang2017}, DARTEL~\cite{Ashburner07}, and SyN~\cite{Avants08}. Unfortunately, these methods often demand substantial time and computational resources to run for a given image pair.

Recent methods have proposed to train neural networks that map a pair of input images to an output deformation. These approaches usually require ground truth registration fields, often derived via more conventional registration tools, which can introduce a bias and necessitate significant preprocessing~\cite{rohe2017,sokooti2017,yang2017}. Some preliminary papers~\cite{devos2017,li2017} explore unsupervised strategies that build on the spatial transformer network~\cite{jaderberg2015}, but are only demonstrated with constrained deformation models such as affine or small displacement fields. Furthermore, they have only been validated on limited volumes, such as 3D patches or 2D slices. 
%
A recent paper avoids these pitfalls, but still does not provide topology-preserving guarantees or probabilistic uncertainty estimates, which yield meaningful information for downstream image analysis~\cite{balakrishnan2018}.



In this paper we present a formulation for registration as conducting variational inference on a probabilistic generative model. This framework naturally results in a learning algorithm that uses a convolutional neural network with an intuitive cost function.  We introduce novel 
\textit{diffeomorphic integration} layers combined with a transform layer to enable unsupervised end-to-end learning for diffeomorphic registration. We present extensive experiments, demonstrating that our algorithm achieves state of the art registration accuracy while providing diffeomorphic deformations, fast runtime and estimates of registration uncertainty. 

\subsection{Diffeomorphic Registration}

Although the method presented in this paper applies to a multitude of deformable representations, we choose to work with diffeomorphisms, and in particular with a stationary velocity field representation~\cite{Ashburner07}.  Diffeomorphic deformations are differentiable and invertible, and thus preserve topology. Let $\bphi: R^3 \rightarrow R^3$ represent the deformation that maps the coordinates from one image to coordinates in another image. 
In our implementation,  the deformation field is defined through the following ordinary differential equation (ODE): 
\begin{equation}
\frac{\partial\bphi^{(t)}}{\partial t} = \bv(\bphi^{(t)})
\label{eq:ode}
\end{equation}
%
%
where $\bphi^{(0)} = Id$ is the identity transformation and~$t$ is time. We integrate  the stationary velocity field $\bv$ over $t=[0,1]$ to obtain the final registration field $\bphi^{(1)}$. 


We compute the integration numerically using \emph{scaling and squaring}~\cite{arsigny2006}, which we briefly review here.
The integration of a stationary ODE represents a one-parameter subgroup of diffeomorphisms.  In group theory, $\bv$ is a member of the Lie algebra and is exponentiated to produce $\bphi^{(1)}$, which is a member of the Lie group: $\bphi^{(1)} =\text{exp}(\bv)$.
  From the properties of one-parameter subgroups, for any scalars $t$ and $t'$, \mbox{$\exp((t+t')\bv) = \exp(t\bv) \circ \exp(t'\bv)$}, where $\circ$ is a composition map associated with the Lie group. 
Starting from~$\bphi^{(1/2^T)} =\bp + \bv(\bp)$ where~$\bp$ is a map of spatial locations, we use the recurrence~$\bphi^{(1/2^{t-1})} = \bphi^{(1/2^t)} \circ \bphi^{(1/2^t)}$ to obtain~$\bphi^{(1)} = \bphi^{(1/2)} \circ \bphi^{(1/2)}$. $T$ is chosen so that $\bv \approx 0$.


%% file: model.tex
\section{Methods}
\label{sec:model}

Let~$\bmoving$ and~$\bfixed$ be 3D images, such as MRI volumes,   
%
%
and let~$\bz$ be a latent variable that parametrizes a transformation function~$\bphi_{\bz}: R^3 \rightarrow R^3$. We use a generative model to describe the formation of~$\bmoving$ by warping~$\bfixed$ into~$\bfixed \circ \bphi_{\bz}$. 
%
%
We propose a variational inference method that uses a neural network of convolutions, diffeomorphic integration, and spatial transform layers. We learn the network parameters in an unsupervised fashion, i.e., without access to ground truth registrations. We describe how the network yields fast diffeomorphic registration of a new image pair~$\bmoving$ and~$\bfixed$, while providing uncertainty estimates.

\vspace{-0.1cm}
\subsection{Generative Model}

We model the prior probability of~$\bz$ as:
\begin{align}
p(\bz) = \mathcal{N}(\bz; \boldsymbol{0}, \bSigma_z),  
\end{align}
where~$\mathcal{N}(\cdot;\bmu,\bSigma)$ is the multivariate normal distribution with mean~$\bmu$ and covariance~$\bSigma$. 
Our work applies to a wide range of representations~$\bz$. For example,~$\bz$ could be a low-dimensional embedding of a dense displacement field, or even the displacement field itself. In this paper, we let~$\bz$ be a stationary velocity field that specifies a diffeomorphism through the ODE~\eqref{eq:ode}. We let~$\bL = \bD - \bA$ be the Laplacian of a neighborhood graph defined on a voxel grid, where~$\bD$ is the graph degree matrix, and~$\bA$ is a voxel neighbourhood adjacency matrix. We encourage spatial smoothness of~$\bz$ by letting~$\bSigma_z^{-1} = \bLambda_z = \lambda \bL$, where~$\bLambda_z$ is a precision matrix and~$\lambda$ denotes a parameter controlling the scale of the velocity field~$\bz$.


We let~$\bmoving$ be a noisy observation of warped image~$\bfixed$:
\begin{align}
	p(\bmoving | \bz ; \bfixed) &= \mathcal{N}(\bmoving ; \bfixed \circ \bphi_{\bz}, \sigma^2\boldsymbol{\mathbbm{I}}),  
	\label{eq:likelihood}
\end{align}
where~$\sigma^2$ reflects the variance of additive image noise.

We aim to estimate the posterior registration probability~$p(\bz|\bmoving;\bfixed)$. Using this, we can obtain the most likely registration field~$\bphi_{\bz}$ for a new image pair~$(\bmoving, \bfixed)$ via MAP estimation, along with an estimate of uncertainty for this registration. 

\vspace{-0.1cm}
\subsection{Learning}

With our assumptions, computing the posterior probability~$p(\bz|\bmoving; \bfixed)$ is intractable. We use a variational approach, and introduce an approximate posterior probability~$q_{\bpsi}(\bz|\bmoving; \bfixed)$ parametrized by~$\bpsi$. We minimize the KL divergence
%
\begin{align}
\min_{\psi} \KL &\left[q_{\bpsi}(\bz|\bmoving ; \bfixed) || p(\bz|\bmoving ; \bfixed)  \right] \nonumber \\
&= \min_{\psi} \Expect_{q} \left[ \log q_{\bpsi}(\bz|\bmoving ; \bfixed) - \log p(\bz|\bmoving ; \bfixed) \right] \nonumber \\
&= \min_{\psi} \Expect_{q} \left[ \log q_{\bpsi}(\bz|\bmoving ; \bfixed) - \log p(\bz, \bmoving, \bfixed) \right] + \log p(\bmoving ; \bfixed) \nonumber \\
&= \min_{\psi} \KL \left[  q_{\bpsi}(\bz|\bmoving; \bfixed) ||  p(\bz)  \right] - \Expect_{q} \left[ \log p(\bmoving | \bz ; \bfixed) \right],
\label{eq:VLB}
\end{align}
%
%
%
%
which is the negative of the \textit{variational lower bound} of the model evidence~\cite{kingma2013}.
%
%
We model the approximate posterior~$q_{\bpsi}(\bz | \bmoving ; \bfixed)$ as a multivariate normal:
\begin{align}
q_{\bpsi}(\bz | \bmoving ; \bfixed) = \mathcal{N}(\bz ; \bmu_{z | \moving, \fixed}, \bSigma_{z |\moving, \fixed}),
\end{align}
where~$\bSigma_{z | \moving, \fixed}$ is diagonal. 

We estimate~$\bmu_{z | \moving, \fixed}$, and~$\bSigma_{z | \moving, \fixed}$ using a convolutional neural network~$\text{def}_{\bpsi}(\bmoving,\bfixed)$ parameterized by~$\bpsi$, as described below.
%
We can therefore learn the parameters~$\bpsi$ by optimizing the variational lower bound~\eqref{eq:VLB} using stochastic gradient methods. Specifically, for each image pair~$\{\bmoving, \bfixed\}$ and samples~$\bz_k\sim q_{\psi}(\bz|\bmoving; \bfixed)$, we can compute~$\bfixed \circ \bphi_{z_k}$,
with the resulting loss:
\begin{align}
&\mathcal{L}(\bpsi; \bmoving, \bfixed) \nonumber\\
 &= - \Expect_{q} \left[ \log p(\bmoving | \bz ; \bfixed) \right] + \KL \left[  q_{\bpsi}(\bz|\bmoving; \bfixed) ||  p(\bz)  \right]  \label{eq:main_loss}\\
&= \frac{1}{2\sigma^2K} \sum_k ||\bmoving - \bfixed \circ \bphi_{z_k}||^2  + \frac{1}{2} \left[ \text{tr}(\lambda\bD \bSigma_{z|x;y} - \log|\bSigma_{z|x;y}|) + \bmu_{z | \moving, \fixed}^T \bLambda_z \bmu_{z | \moving, \fixed} \right] + \text{const}, \nonumber 
\end{align}
where~$K$ is the number of samples used. In our experiments, we use~$K=1$.
The first term encourages the warped image~$\bfixed \circ \bphi_{z_k}$ to be similar to~$\bmoving$. The second term encourages the posterior to be close to the prior~$p(\bz)$. Although the variational covariance~$\bSigma_{z|\moving,\fixed}$ is diagonal, the last term spatially smoothes the mean:~$\bmu_{z | \moving, \fixed}^T \bLambda_z \bmu_{z | \moving, \fixed} = \frac{\lambda}{2} \sum \sum_{j\in N(I)} (\bmu[i] - \bmu[j])^2$, where
%
$N(i)$ are the neighbors of voxel~$i$. 
We treat~$\sigma^2$ and~$\lambda$ as fixed hyper-parameters.

\vspace{-0.2cm}
\subsection{Neural Network Framework}
\vspace{-0.1cm}

\begin{figure}[t]
	\centering
	\begin{minipage}[t]{1\linewidth}
		\includegraphics[width=1\linewidth]{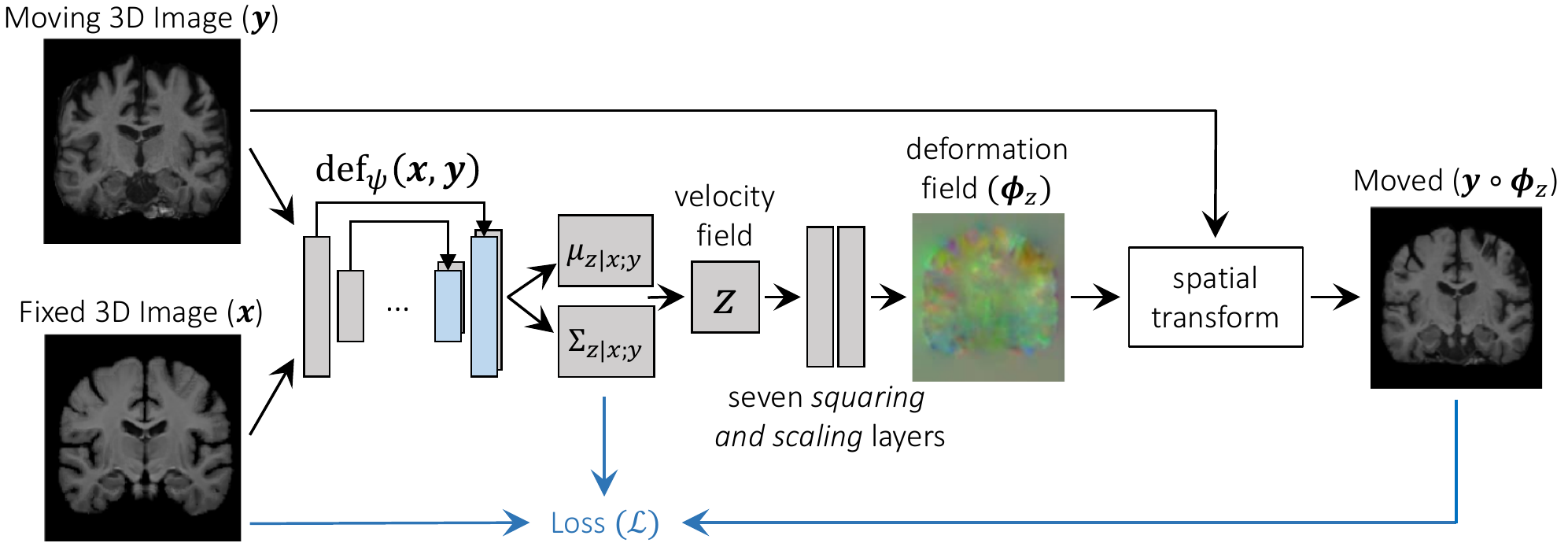}
	\end{minipage}
	\hfill
	%
	\begin{minipage}[b]{1\linewidth}
		\caption{Overview of end-to-end unsupervised architecture. The first part of the network,~$\text{def}_{\psi}(\bmoving, \bfixed)$ takes the input images and outputs the approximate posterior probability parameters representing the velocity field mean,~$\bmu_{z|\moving;\fixed}$, and variance,~$\bSigma_{z|\moving;\fixed}$. A velocity field~$\bz$ is sampled and transformed to a diffeomorphic deformation field~$\bphi_z$ using novel differentiable \textit{squaring and scaling} layers. Finally, a spatial transform warps~$\bfixed$ to obtain~$\bfixed \circ \bphi_z$.
			}
		\label{fig:network_overview_simple}
	\end{minipage}
\end{figure}

We design the network~~$\text{def}_{\bpsi}(\bmoving,\bfixed)$ that takes as input~$\bmoving$ and~$\bfixed$ and outputs~$\bmu_{z|\moving,\fixed}$ and~$\bSigma_{z|\moving,\fixed}$, based on a 3D UNet-style architecture~\cite{ronneberger2015}. 
The network includes a convolutional layer with 16 filters, four downsampling layers with 32 convolutional filters and a stride of two, and finally three upsampling convolutional layers with 32 filters. All convolutional layers use LeakyReLu activations and a 3x3 kernel.

To enable unsupervised learning of parameters~$\bpsi$ using~\eqref{eq:main_loss}, we must form~\mbox{$\by \circ \bphi_z$} to compute the data term. We first implement a layer that samples a new \mbox{$\bz_k \sim \mathcal{N}(\bmu_{z|\moving,\fixed}, \bSigma_{z|\moving,\fixed})$} using the ``re-parameterization trick"~\cite{kingma2013}. 

We propose novel \emph{scaling and squaring} network layers to compute~\mbox{$\bphi_{z_k} = \exp(\bz_k)$}.  
%
%
Specifically, these involve compositions within the neural network architecture using a differentiable spatial transformation operation. Given two 3D vector fields $\ba$ and $\bb$, for each voxel $p$ such a layer computes $(\ba \circ \bb)(p) = \ba(\bb(p))$, a non-integer voxel location $\bb(\bp)$ in $\ba$, using linear interpolation. Starting with~$\bv^{(1/2^T)} = \bz_k$, we compute~\mbox{$\bv^{(1/2^{t+1})} = \bv^{(1/2^t)} \circ \bv^{(1/2^t)}$} recursively using these layers, leading to~$\bv^{(1)} \triangleq \bphi_{z_k} = \exp(\bz_k)$. In our experiments, we use~$T=7$.

Finally, we use a spatial transform layer to warp volume~$\bfixed$ according to the computed diffeomorphic field~$\bphi_{z_k}$. This network results in three outputs,~$\bmu_{z|\moving,\fixed}, \bSigma_{z|\moving,\fixed}$ and~~$\by \circ \bphi_{z_k}$, which are used in the model loss~\eqref{eq:main_loss}.

In summary, the neural network takes as input~$\bmoving$ and~$\bfixed$, computes~$\bmu_{z|\moving,\fixed}$ and~$\bSigma_{z|\moving,\fixed}$, samples a new~$\bz_k \sim \mathcal{N}(\bmu_k, \bSigma_k)$, computes a diffeomorphic~$\bphi_{z_k}$ and applies it to~$\bfixed$. Since all the steps are designed to be differentiable, we  learn the network parameters using stochastic gradient descent based methods on the loss~\eqref{eq:main_loss}. The framework is summarized in Figure~\ref{fig:network_overview_simple}. 

Our implementation uses Keras~\cite{chollet2015} 
 with a Tensorflow backend~\cite{abadi2016}
 and the ADAM optimizer~\cite{kingma2014}. 
 We implement our method as part of the VoxelMorph package, with both implementations available online at \url{https://github.com/voxelmorph/voxelmorph}.

\subsection{Registration and Uncertainty}

Given learned parameters, we approximate registration of a new scan pair~$(\bmoving, \bfixed)$ using~$\bphi_{\hat{z}_k}$. We first obtain~$\hat{\bz}_k$ using
\begin{align}
\hat{\bz}_k &= \arg \max_{\bz_k} p(\bz_k | \bmoving; \bfixed) = \bmu_{z|\moving;\fixed},
\label{eq:MAP1}
\end{align}
by evaluating the neural network~$\text{def}_\psi(\bmoving, \bfixed)$ on the two input images. We then compute~$\bphi_{\hat{z}_k}$ using the \textit{scaling and squaring} network layers. 
We also obtain~$\bSigma_{z|\moving,\fixed}$, enabling an estimation of the uncertainty of the velocity field $\bz$ at each voxel~$j$:
\begin{align}
H(\bz[j]) &\approx \Expect \left[ -\log q_{\bpsi}(\bz|\bmoving,\bfixed) \right] = \frac{1}{2} \log 2 \pi \bSigma_{z|\moving;\fixed}[j,j].
\label{eq:uncertainty_z}
\end{align}
We also estimate uncertainty in the deformation field~$\bphi_z$ empirically. We sample several representations~$\bz_{k'} \sim q_{\psi}(\bz | \bmoving; \bfixed)$, propagate them through the diffeomorphic layers to compute~$\bphi_{z_k'}$, and compute the empirical diagonal covariance~$\hat{\bSigma}_{\bphi_z}[j,j]$ across samples. The uncertainty is then~$H(\bphi[j]) \approx \frac{1}{2} \log 2 \pi \hat{\bSigma}_{\phi_z}[j,j]$.
%


%% file: results.tex
\section{Experiments}
\label{sec:results}

\begin{figure}[b!]
	\begin{center}
		\includegraphics[width=\linewidth]{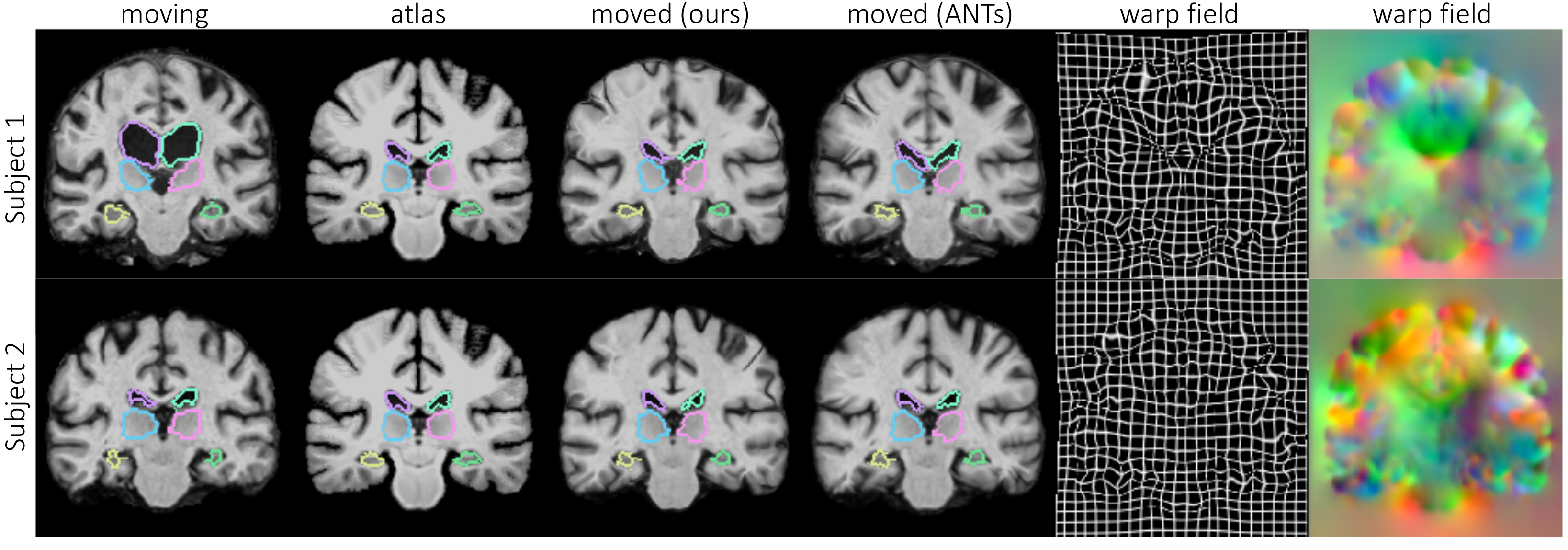}
	\end{center}
	\vspace{-0.5cm}
	\caption{ Example MR slices of input moving image, atlas, and resulting warped image for our method and ANTs, with overlaid boundaries of ventricles, thalami and hippocampi. Our resulting registration field is shown as a warped grid and RGB image, with each channel representing dimension. Due to space constraints, we omit VoxelMorph examples, which are \textit{visually} similar to our results and ANTs.}
	\label{fig:reg_examples}
\end{figure}

We focus on 3D atlas-based registration, a common task in population analysis. Specifically, we register each scan to an atlas computed using external data~\cite{fischl2012}. 

\vspace{0.3cm}
\noindent \textbf{Data and Preprocessing.} We use a large-scale, multi-site dataset of 7829 T1-weighted brain MRI scans from eight publicly available datasets: ADNI~\cite{mueller2005ways}, OASIS~\cite{marcus2007open}, ABIDE~\cite{di2014autism}, ADHD200~\cite{milham2012adhd}, MCIC~\cite{gollub2013mcic}, PPMI~\cite{marek2011parkinson}, HABS~\cite{dagley2015harvard}, and Harvard GSP~\cite{holmes2015brain}. Acquisition details, subject age ranges and health conditions are different for each dataset. We performed standard pre-processing steps on all scans, including resampling to~$1$mm isotropic voxels, affine spatial normalization and brain extraction for each scan using FreeSurfer~\cite{fischl2012}. We crop the final images to $160\times 192 \times 224$. Segmentation maps including 29 anatomical structures, obtained using FreeSurfer for each scan, are used in evaluating registration results. We split the dataset into 7329, 250, and 250 volumes for train, validation, and test sets respectively, although we underscore that the training is unsupervised.

\begin{table}[t!]
	\small
	\centering
	\begin{tabular}{c c c c c c}
		Method&Avg. Dice&GPU sec&CPU sec & $|J_\Phi| \le 0$ & Uncertainty   \\
		\hline
		Affine only&0.567 (0.157)&0 &0  & 0 & No \\
		ANTs (SyN) &0.750 (0.135)&-&9059 (2023) & 6505 (3024) & No \\
		VoxelMorph & 0.750 (0.137)&0.554 (0.017) & 144 (1) & 18096 (4230) & No   \\
		\textbf{Ours} & \textbf{0.753} (0.137) & \textbf{0.451} (0.011)& \textbf{51} (0.2) & \textbf{0.7} (2.0) & \textbf{Yes}  \\
		\hline  
	\end{tabular}
	\vspace{+0.2cm}
	\caption{ Summary of results: mean Dice scores over all anatomical structures and subjects (higher is better), mean runtime; and mean number of locations with a non-positive Jacobian of each registration field (lower is better). All methods have comparable Dice scores, while our method and the original VoxelMorph are orders of magnitude faster than ANTs. Only our method
		achieves both high accuracy and fast runtime while also having nearly zero non-negative Jacobian locations and providing uncertainty prediction.
		 	\vspace{-0.5cm}
		 	}
	\label{tbl:results}
\end{table}   
\normalsize


	\vspace{0.1cm}\noindent \textbf{Evaluation Metric.}  To evaluate a registration algorithm, we register each subject to an atlas, propagate the segmentation map using the resulting warp, and measure volume overlap using the Dice metric. We also evaluate the diffeomorphic property, a focus of our model. Specifically, the Jacobian matrix \mbox{$J_{\phi}(p) = \nabla \bphi (p) \in \mathcal{R}^{3\times 3}$} captures the local properties of $\bphi$ around voxel $p$. The local deformation is diffeomorphic, both invertible and orientation-preserving, only at locations for which $|J_{\phi}(p)| > 0$~\cite{Ashburner07}.
	We count all other voxels, where~$|J_{\phi}(p)| \le 0$.
	
	
	\begin{figure*}[b!]
		\begin{center}
			\includegraphics[width=\linewidth]{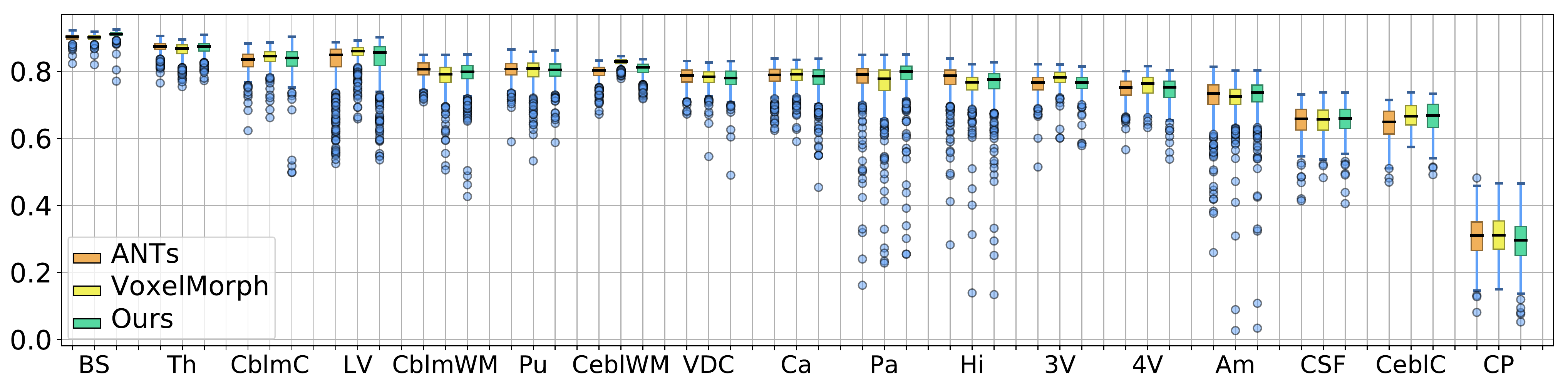}
		\end{center}
		\vspace{-0.4cm}
		\caption{Boxplots indicating Dice scores for anatomical structures for ANTs, VoxelMorph, and our algorithm. 
	 Left and right hemisphere structures are merged for visualization, and ordered by average ANTs Dice score. We include the brain stem (BS), thalamus (Th), cerebellum cortex (CblmC), lateral ventricle (LV), cerebellum white matter (CblmWM), putamen (Pu), cerebral white matter (CeblWM), ventral DC (VDC), caudate (Ca), pallidum (Pa), hippocampus (Hi), 3rd ventricle (3V), 4th ventricle (4V), amygdala (Am), CSF (CSF), cerebral cortex (CeblC), and choroid plexus~(CP). 
		}
		\label{fig:boxplot}
	\end{figure*}

	\vspace{0.2cm}\noindent \textbf{Baseline Methods.} We compare our approach with the popular ANTs software package using Symmetric Normalization (SyN)~\cite{Avants08}, a top-performing algorithm~\cite{klein2009}.  We found that the default ANTs settings were sub-optimal for our task, so we performed a wide parameter and similarity metric search across a multitude of datasets. We identified top performing parameter values on the Dice metric and used cross-correlation as the ANTs similarity measure. We also test our recent CNN-based method, VoxelMorph, which aims to produce fast registration but does not yield diffeomorphic results or uncertainty estimates~\cite{balakrishnan2018}. We sweep the regularization parameter using our validation set, and use the optimal parameters in our results.


	\vspace{0.2cm}\noindent \textbf{Results on Test set:} Figure~\ref{fig:reg_examples} shows  representative results.  Figure~\ref{fig:boxplot} illustrates Dice results on a range of anatomical structures, and Table~\ref{tbl:results} gives a summary of the results.  Not only does our algorithm achieve state of the art Dice results and the fastest runtimes, but it also produces diffeomorphic registration fields (having nearly no non-negative Jacobian voxels per scan) and yields uncertainty estimates. 
	
	Specifically, all methods achieve comparable Dice results on each structure and overall. Our method and VoxelMorph require a fraction of the ANTs runtime to register two images: less than a second on a GPU, and less than a minute on a CPU (for our method). To the best of our knowledge, ANTs does not have a GPU implementation. Algorithm runtimes were computed for an NVIDIA TitanX GPU and a Intel Xeon (E5-2680) CPU, and exclude preprocessing common to all methods. Importantly, while our method achieves positive Jacobians at nearly all voxels, the flow fields resulting from the baseline methods contain a few thousand locations of non-positive Jacobians. This can be alleviated with increased spatial regularization, but this in turn leads to a drop in performance on the Dice metric.

%



\begin{figure}[t]
	\centering
	\includegraphics[width=1\linewidth]{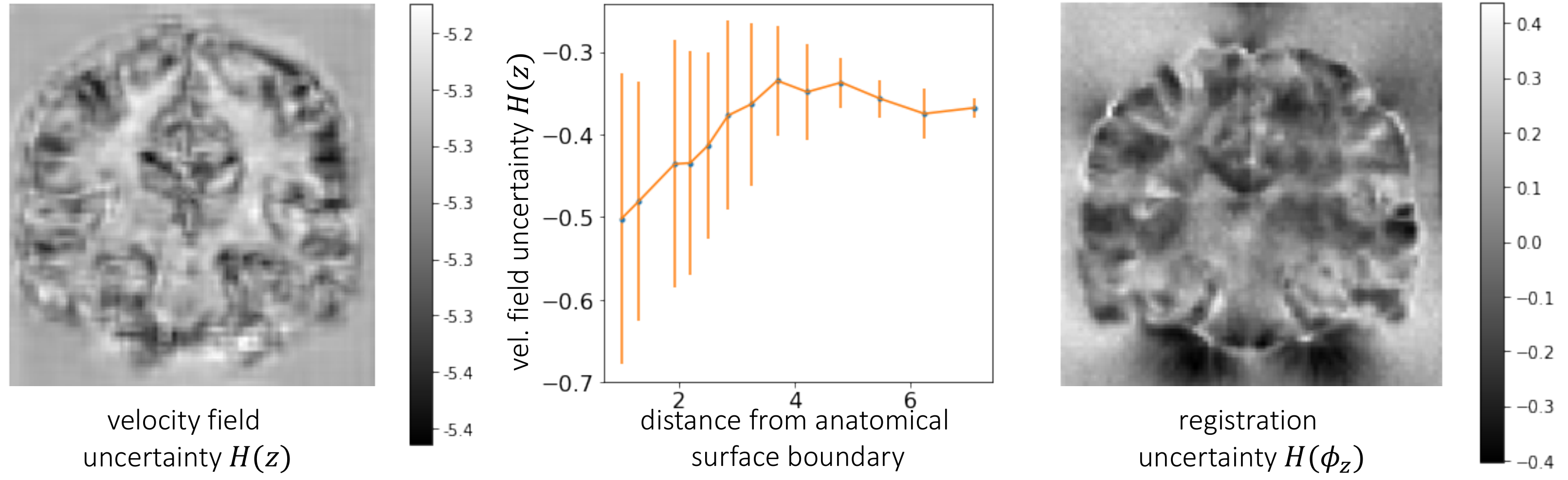}
	%
		\caption{\scriptsize Example velocity field uncertainty~$H(\bz)$ (left) indicates low uncertainty near structure boundaries, as seen in the line graph (middle). 
		This correlation is less obvious in the final registration field uncertainty~$H(\bphi_z)$ (right).
		}
		\label{fig:uncertainty}
\end{figure}

\vspace{0.2cm}\noindent \textbf{Uncertainty.}	Figure~\ref{fig:uncertainty} shows representative uncertainty maps, unique to our model. The velocity field is more certain near anatomical structure edges, and less confident in homogenous scan regions, such as the white matter or ventricle interior. 

\vspace{0.2cm}\noindent \textbf{Parameter Analysis.} We perform a grid search for the two fixed hyper-parameters~$\lambda$ and~$\sigma^2$. We train a model for each parameter pair and evaluate Dice on the validation set. We search 30 values within two orders of magnitude around meaningful initial values for both parameters:~$\sigma^2 \sim (0.07)^2$, the variance of the intensity difference between an \textit{affinely} aligned image and the atlas, and~$\lambda = 10000$, equivalent to a diagonal standard deviation of 1 voxel for~$\bphi_z$. We found~$\sigma^2 \sim (0.035)^2$ and~$\lambda \in (20000, 100000)$ to perform well, and set~$\lambda=70,000$.



%% file: conclusion.tex
\section{Conclusion}

We propose a probabilistic model for diffeomorphic image registration and derive a learning algorithm that makes use of a convolutional neural network and an intuitive resulting loss function. To achieve unsupervised, end-to-end learning for diffeomorphic registrations, we introduce novel \textit{scaling and squaring} differentiable layers. Our derivation is generalizable. For example,~$\bz$ can be a low dimensional embedding representation of a deformation field, or the displacement field itself. 
Our algorithm can infer the registration of new image pairs in under a second. Compared to traditional methods, our method is significantly faster, and compared to recent learning based methods, our method offers diffeomorphic guarantees, and provides natural uncertainty estimates for resulting registrations.

\section{Acknowledgments}
This research was funded, in part, by NIH grants R01LM012719, R01AG053949, and 1R21AG050122, and NSF grant 1707312, the Cornell NeuroNex Hub. 